\documentclass{article}

\usepackage[preprint]{neurips_2026}

\usepackage[utf8]{inputenc} 
\usepackage[T1]{fontenc}   
\usepackage{hyperref}      
\usepackage{url}            
\usepackage{booktabs}       
\usepackage{graphicx}       
\usepackage{amsfonts}       
\usepackage{nicefrac}       
\usepackage[nopatch=footnote]{microtype}      
\usepackage{xcolor}        
\usepackage{wrapfig}
\usepackage{caption}
\usepackage{tcolorbox} 
\tcbuselibrary{breakable} 
\usepackage{setspace}
\usepackage{pifont}
\newcommand{\cmark}{\ding{51}}
\newcommand{\xmark}{\ding{55}}  
\usepackage{colortbl}
\usepackage{enumitem}
\usepackage{amsmath}
\usepackage{cleveref}
\usepackage{subcaption}
\usepackage{multirow}
\usepackage{arydshln}

\title{ALIVE: Awakening LLM Reasoning via Adversarial Learning and Instructive Verbal Evaluation}

\author{%
Yiwen Duan \thanks{Equal contribution.} \\
Independent Researcher \\
\texttt{berryallen.usa@gmail.com} \\
\And
Jing Ye \footnotemark[1]\hspace{0.2em} \thanks{Corresponding author.} \\
Independent Researcher \\
\texttt{jing.ye678@gmail.com} \\
\And
Xinpei Zhao \\
Independent Researcher \\
\texttt{victor.ma.zhao@gmail.com} \\
}

\begin{document}

\maketitle

\begin{abstract}
Achieving expert-level reasoning in Large Language Models (LLMs) remains constrained by a persistent reward bottleneck: conventional reinforcement learning (RL) relies on outcome scalar rewards that are \textbf{costly} to scale, \textbf{brittle} across domains, and \textbf{blind} to the underlying logic of a solution. This reliance on impoverished external signals limits the model’s ability to develop a deep, self-contained understanding of reasoning principles.
We propose to view reasoning as a \textbf{self-evolving process}, in which learning arises from a closed-loop interaction among problem construction, solution generation, and review.
Therefore, we introduce \textbf{ALIVE} (\textbf{A}dversarial \textbf{L}earning with \textbf{I}nstructive \textbf{V}erbal \textbf{E}valuation), a unified self-supervised alignment framework that shifts the objective from reward optimization to intrinsic reasoning acquisition.
Specifically, ALIVE organizes learning as a closed-loop interaction among three roles: task construction, problem solving, and solution review. Starting from raw corpora, the model first constructs reasoning tasks by masking informative spans and generating corresponding targets, then produces reasoning trajectories to solve them, and finally critiques its own solutions through instructive verbal evaluations that assess both correctness and reasoning quality. This adversarial self-review process converts raw text into scalable reasoning supervision, enabling the model to iteratively refine its reasoning behavior without externally annotated rewards.
Across benchmarks in mathematical reasoning, code generation, and logical inference, ALIVE consistently outperforms strong baselines under matched data and compute budgets, achieving higher accuracy, improved cross-domain generalization, and stronger self-correction ability. These results suggest that closed-loop reasoning acquisition provides a scalable alternative to reward-driven learning, positioning ALIVE as a step toward self-evolving reasoning systems.
\end{abstract}

\section{Introduction}
Large Language Models (LLMs) have achieved substantial progress, advancing from core language understanding \citep{GPT3, GPT} to complex reasoning tasks such as mathematical problem solving \citep{DeepSeekMath} and code generation \citep{qwen25-coder}. These gains are largely driven by two factors: scaling data and model capacity \citep{scalinglaws, NEURIPS2022_c1e2faff} and reinforcement learning (RL)-based post-training \citep{DeepSeek-R1, openai2024openaio1card}, where models are optimized using external reward signals. While effective, this paradigm treats reasoning as a byproduct of reward maximization, rather than as an ability that can be learned and refined in its own right.

In this work, we argue that this paradigm is fundamentally limited by a persistent \textbf{reward bottleneck}:

\paragraph{1. Costly to scale.} RLHF \citep{RLHF} depends on expensive, noisy human feedback, while RLAIF \citep{RLAIF} scales with synthetic rules but often introduces biases that misalign with real-world reasoning \citep{rlhfbook}. RLVR \citep{Tulu3, jimenez2024swebenchlanguagemodelsresolve} is restricted to easily constructed environments, limiting its use in realistic, open-ended tasks due to the high cost and complexity of building such settings.

\paragraph{2. Brittle across domains.} Reward models typically exploit superficial correlations rather than fundamental reasoning \citep{wang2025reinforcementlearningenhancedllms}, requiring labor-intensive, task-specific reward engineering \citep{wang2024secretsrlhflargelanguage, mahan2024generativerewardmodels}. Consequently, models trained in one domain rarely generalize to other domains, resulting in brittle performance across diverse tasks.

\paragraph{3. Blind to the underlying logic of the reward.} Scalar or binary rewards discard the semantic structure of multi-step reasoning \citep{feng2024naturallanguagereinforcementlearning, liu2025spiceselfplaycorpusenvironments, Text2Grad, lightman2023let, mukherjee2023orca, FCP}, forcing models to rely on trial-and-error. Process-based rewards, chain-of-thought supervision, and verbal feedback help but still depend on curated data or human oversight, leaving the reward bottleneck largely unresolved.

More fundamentally, reliance on externally specified rewards prevents the emergence of a closed-loop learning dynamic. This raises a central question:

\begin{tcolorbox}[breakable]
\footnotesize
\setstretch{1}
\textit{Can a language model evolve its reasoning ability through a self-contained learning process, where tasks, solutions, and evaluative signals are generated intrinsically rather than externally specified?}
\end{tcolorbox}

We propose to reframe LLM reasoning as a \textbf{self-evolving system}, in which learning is driven by the interaction between problem construction, solution generation, and self-evaluation. Under this perspective, reasoning ability is not induced by reward signals, but emerges from the model’s ability to iteratively refine its own reasoning process. To instantiate this paradigm, we introduce \textbf{ALIVE}, a unified self-supervised framework that integrates three roles—task construction, problem solving, and solution reviewing—within a single policy.  ALIVE forms a closed-loop adversarial process that internalizes reasoning correctness from raw text alone, eliminating the need for external reward annotation. As shown in Figure~\ref{fig:alive-framework}, ALIVE operates in three stages: \textbf{(1) Task Construction:} The model autonomously constructs reasoning tasks by masking valuable spans in raw text and generating corresponding ground-truth targets, creating a scalable and domain-agnostic source of supervision. \textbf{(2) Problem Solving:} The model produces complete reasoning trajectories and predicts the masked spans. \textbf{(3) Solution Review:} The model evaluates its own predictions using natural language critiques that explicitly assess correctness and reasoning quality. ALIVE circumvents the reward bottleneck by using self-generated, reasoning-rich verbal critiques derived from raw text as scalable, information-dense supervision.

Empirical evaluations across benchmarks for mathematical reasoning, code generation, and general logical inference demonstrate that ALIVE effectively alleviates the reward bottleneck. Under identical data and compute budgets, ALIVE delivers substantial performance gains, exhibits stronger cross-domain generalization, and achieves markedly higher self-correction rates. Collectively, these results indicate that the proposed reasoning trinity enables a self-reinforcing trajectory of capability growth, positioning ALIVE as a scalable foundation for general-purpose reasoning alignment that relies on no human-in-the-loop supervision.

Our main contributions are summarized as follows:
\begin{itemize}[itemsep=0.1pt, topsep=0.1pt, partopsep=0.1pt, leftmargin=*]
\item We formally identify and systematize the \textbf{reward bottleneck} in RL-based reasoning, uncovering its fundamental limitations in scalability, robustness, and information efficiency.
\item We introduce \textbf{ALIVE}, a unified self-supervised RL framework that enables LLMs to autonomously construct, solve, and review reasoning tasks directly from raw text.
\item We provide broad empirical evidence that ALIVE outperforms scalar-reward, verbal-feedback, self-distillation, and self-evolving baselines under matched data and compute, while role-level ablations validate the necessity of the three-role training loop.

\end{itemize}

\section{The ALIVE Framework}
\label{sec:method}

The ALIVE framework is motivated by a fundamental question: if external rewards are costly, brittle, and weakly informative, can a language model autonomously construct its own training supervision directly from raw text? Our approach transforms each document into a self-contained, closed-loop training environment. Within this environment, the model first masks a meaningful text span, attempts to infer it from the remaining context, and subsequently reviews its generated reasoning trajectory against the hidden span. The masked span serves as an automatic, objective ground truth, while the subsequent review translates this ground truth into both a scalar learning signal and a natural-language feedback.

\begin{figure}[t]
    \centering
    \includegraphics[width=\textwidth]{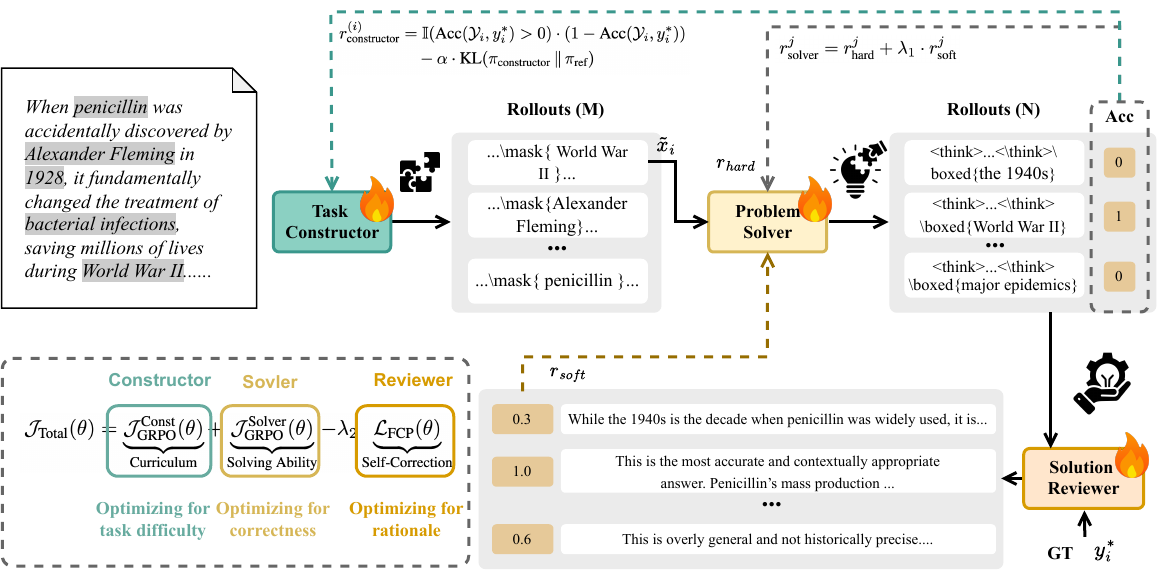}
    \caption{\textbf{Overview of ALIVE.} A unified policy model $\pi_\theta$ alternates among three roles in a closed-loop learning cycle: the \textbf{Constructor} masks reasoning-critical spans in text to create hindsight-verifiable tasks, the \textbf{Solver} generates reasoning trajectories for these tasks, and the \textbf{Reviewer} evaluates the resulting solutions with both verbal critiques and soft rewards. The same parameters are updated by combining task-difficulty, hard-verification, soft-review, and feedback-conditioned learning signals.}
    \label{fig:alive-framework}
\end{figure}

A single, unified policy $\pi_\theta$ executes all three phases of this loop. By employing role-specific prompts, we condition the model to act as a \textit{Constructor}, \textit{Solver}, or \textit{Reviewer} while keeping its underlying parameters fully shared. This parameter sharing is central to ALIVE: the policy that learns to solve reasoning tasks simultaneously learns which tasks are structurally useful and what constitutes a valid explanation of success or failure. Figure~\ref{fig:alive-framework} provides a schematic overview of this framework.

\subsection{Role-Conditioned Closed-Loop Reasoning}
\label{sec:roles}
Let $\mathcal{D}=\{d\}$ denote a textual corpus, where each document $d$ can be either a natural language text or a tokenized sequence converted from a structured data source. Rather than relying on human-annotated rewards, ALIVE extracts supervision from masked spans within the corpus. For each document, the unified policy sequentially alternates among three distinct roles:

\textbf{\raisebox{-0.3em}{\includegraphics[height=1.5em]{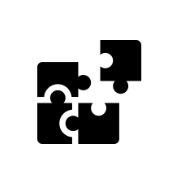}} The Constructor.}
Given a document $d$, the policy acts as a stochastic task generator. It performs $M$ independent construction rollouts. Each rollout selectively masks a reasoning-critical span, yielding a query $\tilde{x}_i$ from the remaining context. The omitted span serves as the \textit{Hindsight Ground Truth} $y_i^*$:
\begin{equation}
    \mathcal{T}_i
    =
    (\tilde{x}_i,y_i^*)
    \sim
    \pi_\theta( \cdot \mid d).
    \label{eq:1}
\end{equation}
The objective of the generated task is to reconstruct the masked span. This formulation ensures that every auto-generated task is inherently verifiable against a known target, while allowing the masking policy to remain adaptive.

\textbf{\raisebox{-0.3em}{\includegraphics[height=1.5em]{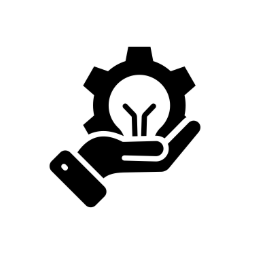}} The Solver.}
For each constructed query $\tilde{x}_i$, the policy samples $N$ candidate solutions. Each solution $\hat{y}_{ij}$ comprises an explicit reasoning trace $z_{ij}$ followed by the final predicted answer $a_{ij}$:
\begin{equation}
    \mathcal{Y}_i=\{\hat{y}_{i1},\ldots,\hat{y}_{iN}\}\sim\pi_\theta(\cdot\mid \tilde{x}_i),\qquad \hat{y}_{ij}=(z_{ij},a_{ij}).
\end{equation}
By treating the intermediate reasoning trace as an integral component of the generative output, downstream evaluations can measure both final answer accuracy and the logical coherence of the solution path.

\textbf{\raisebox{-0.3em}{\includegraphics[height=1.5em]{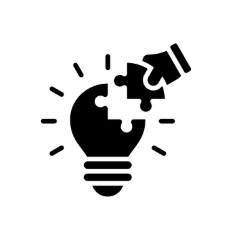}} The Reviewer.}
Finally, the policy $\pi_\theta$ evaluates each candidate solution by conditioning on the query, the generated reasoning-answer pair, and the hindsight ground truth:
\begin{equation}
    (c_{ij},v_{ij})\sim\pi_\theta(\cdot\mid \tilde{x}_i,\hat{y}_{ij},y_i^*).
\end{equation}
Here, $c_{ij}$ represents a detailed, natural-language verbal critique, and $v_{ij}\in[0,1]$ denotes a soft score reflecting partial logical correctness. Crucially, the Reviewer does not function merely as a scalar reward model; it provides a diagnostic explanation of the reward signal, which subsequently serves as a natural-language feedback to guide the policy.

\subsection{Learning to Construct Tasks}
\label{sec:masking}
To maximize sample efficiency, the Constructor must avoid masking trivial or impossible spans. If a span is too easily inferred, all Solver rollouts will recover it successfully, yielding negligible learning signals. Conversely, if a span is decoupled from its surrounding context, the task introduces optimization noise rather than meaningful supervision. Consequently, ALIVE trains the Constructor to generate valid tasks that reside at the frontier of the Solver's current capabilities.

For a given task $i$, we define the empirical exact-match accuracy across the $N$ Solver rollouts as:\begin{equation}
    \mathrm{Acc}(\mathcal{Y}_i,y_i^*)
    =
    \frac{1}{N}
    \sum_{j=1}^{N}
    \mathbb{I}\!\left(\mathrm{ExactMatch}(a_{ij},y_i^*)\right).
    \label{eq:task_accuracy}
\end{equation}
We assign a positive reward to the Constructor only when a task is solved by at least one rollout but remains sufficiently challenging to prevent a perfect success rate across all rollouts:
\begin{equation}
    r_{\mathrm{C}}^{i}
    =
    \mathbb{I}\!\left(\mathrm{Acc}(\mathcal{Y}_i,y_i^*)>0\right)
    \cdot
    \left(1-\mathrm{Acc}(\mathcal{Y}_i,y_i^*)\right).
    \label{eq:constructor_reward}
\end{equation}
The indicator function penalizes unsolvable tasks under the current policy, preventing the Constructor from isolating contexts with insufficient information. Meanwhile, the second term prioritizes tasks that fall near the Solver's current decision boundary, where partial success coexists with substantial uncertainty. This reward design instantiates an intrinsic, adaptive curriculum that continuously tailors task difficulty to the evolving proficiency of the Solver.

We optimize the Constructor using Group Relative Policy Optimization (GRPO)~\citep{DeepSeekMath}. The objective is formulated as:
\begin{equation}
\label{eq:grpo_const}
\begin{gathered}
\mathcal{J}_{\mathrm{C}}(\theta) = \mathbb{E}_{d\sim\mathcal{D}} \left[
\frac{1}{M} \sum_{i=1}^{M} \left( \min\!\left( \rho_i\hat{A}_i, \mathrm{clip}(\rho_i,1-\epsilon,1+\epsilon)\hat{A}_i \right) - \alpha\,\mathbb{D}_{\mathrm{KL}}^{i} \right) \right],
\\
\hat{A}_i
=
\frac{
r_{\mathrm{C}}^{i}
-
\mathrm{mean}(\{r_{\mathrm{C}}^{i}\}_{i=1}^M)
}{
\mathrm{std}(\{r_{\mathrm{C}}^{i}\}_{i=1}^M)
}, 
\quad
\rho_i
=
\frac{
\pi_\theta(s_i\mid d)
}{
\pi_{\theta_{\mathrm{old}}}(s_i\mid d)
}.
\end{gathered}
\end{equation}
$\rho_i$ is the importance-sampling ratio. $\mathbb{D}_{\mathrm{KL}}^{i}$ denote the KL divergence relative to the reference policy. The normalized advantage compares tasks constructed from the same document, encouraging the Constructor to prefer masks that induce more challenging, reasoning-intensive prediction tasks over other locally plausible alternatives.

\subsection{Learning to Solve with Hard and Soft Feedback}
\label{sec:review}

The Solver is trained via two complementary feedback signals. The first is a sparse, binary reward derived directly from the hindsight target:
\begin{equation}
    r_{\mathrm{hard}}^{j}
    =
    \mathbb{I}\!\left(\mathrm{ExactMatch}(a_{ij},y_i^*)\right).
    \label{eq:hard_reward}
\end{equation}
While this hard reward provides an objective anchor for correctness, it suffers from severe sparsity: reasoning trajectories that are logically sound but contain minor surface-level deviations are penalized identically to entirely irrelevant generations.

To provide denser optimization signals, we incorporate a soft reward $r_{\mathrm{soft}}^{j}=v_{ij}$ generated by the Reviewer. Because the Reviewer explicitly conditions on the hindsight ground-truth $y_i^*$, it can assess whether the intermediate reasoning steps are logical, and whether the final answer is semantically consistent with the hidden text. This soft reward explicitly mitigates two common failure modes of exact-match objectives: (1) instances where the reasoning trajectory is entirely correct but the final answer contains minor syntactic variations, and (2) scenarios where the generated answer is contextually plausible yet differs from the exact verbatim span of the original document.

We combine the two signals as:
\begin{equation}
    r_{\mathrm{S}}^{j}
    =
    r_{\mathrm{hard}}^{j}
    +
    \lambda_1 r_{\mathrm{soft}}^{j}.
    \label{eq:solver_reward}
\end{equation}
The hard reward grounds the optimization process to prevent reward hacking and drift, while the soft reward provides a learning signal for partially correct trajectories. Within the $N$ rollouts allocated for a given task, ALIVE normalizes these rewards to compute relative advantages:
\begin{equation}
    A_{ij}
    =
    \frac{
    r_{\mathrm{S}}^{j}
    -
    \mathrm{mean}(\{r_{\mathrm{S}}^{j}\}_{j=1}^N)
    }{
    \mathrm{std}(\{r_{\mathrm{S}}^{j}\}_{j=1}^N)
    }.
    \label{eq:solver_advantage}
\end{equation}

The Solver optimization objective adapts the GRPO:
\begin{equation}
\label{eq:solver_grpo}
\begin{gathered}
\mathcal{J}_{\mathrm{S}}(\theta)
=
\mathbb{E}
\left[
\frac{1}{N}
\sum_{j=1}^{N}
\left(
\min\!\left(
\rho_{ij}A_{ij},
\mathrm{clip}(\rho_{ij},1-\epsilon,1+\epsilon)A_{ij}
\right)
-
\beta\,\mathbb{D}_{\mathrm{KL}}^{ij}
\right)
\right].
\end{gathered}
\end{equation}
where the importance sampling ratio is defined as:
\begin{equation}
    \rho_{ij}
    =
    \frac{
    \pi_\theta(y_{ij}\mid\tilde{x}_i)
    }{
    \pi_{\theta_{\mathrm{old}}}(y_{ij}\mid\tilde{x}_i)
    }.
    \label{eq:solver_ratio}
\end{equation}
As with the Constructor update, the clipping mechanism serves as the primary trust-region constraint. In our experiments, we retain the KL divergence term in the formulation but set its coefficient to $\beta=0$.

\subsection{Learning from Verbal Critiques}
\label{sec:critique_learning}

While scalar rewards dictate the probability scaling of a trajectory, they fail to specify the underlying causal factors of success or failure. ALIVE leverages the natural-language critiques of the Reviewer to make these latent learning criteria explicit. For instance, a critique can articulate specific errors—such as variable misallocation, missing logical steps, or accidental correct answers derived from flawed premises—that are completely lost when compressed into a scalar value.

To exploit this rich feedback, we utilize Feedback Conditional Policy learning (FCP)~\citep{FCP}. Given a critique $c_{ij}$, the policy is trained via standard autoregressive generation to reconstruct its own reasoning-answer path:
\begin{equation}
\label{eq:fcp_loss}
    \mathcal{L}(\theta)
    =
    -
    \frac{1}{MN}
    \sum_{i=1}^{M}
    \sum_{j=1}^{N}
    \log
    \pi_\theta(y_{ij}\mid \tilde{x}_i,c_{ij}).
\end{equation}
This objective acts as a regularizer alongside reinforcement learning; it trains the policy to map specific diagnostic critiques to their corresponding reasoning trajectories. For successful rollouts, the critique reinforces the validity of the logical chain; for failed rollouts, it explicitly isolates the flawed step, equipping the model with a contextual mechanism to recognize and avoid analogous errors in future generations.

\subsection{Overall Objective and Training Procedure}
\label{sec:global_update}

ALIVE unifies the three role-conditioned objectives into a joint optimization function:
\begin{equation}
    \mathcal{J}_{\mathrm{Total}}(\theta)=
    \underbrace{\mathcal{J}_{\text{GRPO}}^{\mathrm{Constructor}}(\theta)}_{\mathrm{Curriculum}}
    +
    \underbrace{\mathcal{J}_{\text{GRPO}}^{\mathrm{Solver}}(\theta)}_{\mathrm{Reasoning}}
    -
    \lambda_2
    \underbrace{\mathcal{L}_{\mathrm{FCP}}^{\mathrm{Reviewer}}(\theta)}_{\mathrm{Self-Correction}}.
    \label{eq:total_objective}
\end{equation}
The same parameters appear in all three terms. Thus, ALIVE does not train a task generator, a solver, and a judge as independent components. It trains one policy whose task distribution, solution behavior, and review criteria are updated together. This coupling is the mechanism by which ALIVE avoids a fixed external reward source. The Constructor changes the tasks as the Solver changes. The Reviewer gives feedback that is grounded by the hidden target but richer than exact match. The FCP loss then feeds the Reviewer's language back into the policy as a condition for reasoning. A training iteration, therefore, produces its own tasks, attempts, evaluations, and corrective signals from raw text.

In practice, we start with a short critique warm-up to calibrate the review format using an external reviewer. After this warm-up, the external reviewer loss is disabled. ALIVE then proceeds with the closed-loop training procedure described above: sample documents, construct hindsight-verifiable tasks, solve them with multiple rollouts, review each rollout, and update the shared policy with Eq.~\eqref{eq:total_objective}.

\section{Experiments}
\label{sec:experiments}

Our experimental evaluation is designed to progressively validate ALIVE as a self-evolving reasoning system. Concretely, we aim to analyze four questions:

\textbf{RQ1:} Does verbal critique provide a richer learning signal than sparse rewards?

\textbf{RQ2:} Can self-supervision match or exceed costly external annotations?

\textbf{RQ3:} Does the curriculum transfer to complex, open-ended domains? 

\textbf{RQ4:} Why must the three roles co-evolve within a unified policy? 

\paragraph{Experimental Configurations.}
We examine two experimental configurations: \textbf{ALIVE-Self}, in which the policy autonomously generates its own verbal critiques and soft rewards without any external supervision, and \textbf{ALIVE-External}, where verbal critiques and soft rewards are supplied by Kimi-K2~\citep{kimiteam2025kimik2openagentic}. For reproducibility, additional implementation details, further analyses, and the full set of prompts are provided in Appendices~\ref{app:training_details}, \ref{app:additional experiment}, and \ref{app:prompts}.
\subsection{RQ1: Rich Feedback in Adversarial Self-Learning}
\label{sec:exp-verbal-feedback}
\begin{table}[t]
\setstretch{1.2}
\centering
\caption{\textbf{Reasoning results after self-evolving training.} All backbones are base models. Some baseline results are taken from \citet{pretrainZero} and \citet{R-Zero}.}
\label{tab:self_evolving_results}
\resizebox{\textwidth}{!}{%
\begin{tabular}{lllllll}
\toprule[1.2pt]
\multicolumn{1}{c}{{\textbf{Method}}} &
  \multicolumn{1}{c}{\textbf{MMLU-Pro}} &
  \multicolumn{1}{c}{\textbf{SuperGPQA}} &
  \multicolumn{1}{c}{\textbf{BBEH}} &
  \multicolumn{1}{c}{\textbf{Math500}} &
  \multicolumn{1}{c}{\textbf{GSM8K}} &
  \multicolumn{1}{c}{\textbf{AIME24}} \\ \midrule
\rowcolor[HTML]{D9F5D6} 
Qwen3-4B-Base        & 51.94          & 26.32          & 8.67           & 73.30          & 86.30          & 0.00           \\
+ Rand.                & 55.21          & 29.10          & 9.45           & 74.80          & 87.50          & 10.00          \\
+ R-Zero               & 55.47          & 27.55          & 10.42          & 79.60          & 92.12          & 13.40          \\
+ PretrainZero              & 60.37          & 32.28          & 12.68          & 79.10          & 92.90          & 13.30          \\ \midrule
+ ALIVE-Self           & 62.79          & 34.39          & 14.00          & \textbf{83.80} & 92.72          & 13.44          \\
+ ALIVE-External         & \textbf{63.15} & \textbf{34.82} & \textbf{14.56} & 81.45          & \textbf{94.09} & \textbf{15.73} \\ \midrule \midrule
\rowcolor[HTML]{D9F5D6} 
Qwen3-8B-Base        & 59.19          & 31.12          & 10.49          & 70.10          & 91.50          & 10.00          \\
+ Rand.                & 61.59          & 34.19          & 12.96          & 79.20          & 93.80          & 13.30          \\
+ R-Zero               & 61.53          & 31.38          & 10.60          & 82.00          & 94.09          & 15.40          \\
+ PretrainZero              & 64.28          & 34.46          & 14.67          & 81.90          & 93.50          & 20.00          \\ \midrule
+ ALIVE-Self           & 66.82          & \textbf{36.90} & 16.35          & 83.80          & \textbf{96.06} & \textbf{24.06} \\
+ ALIVE-External         & \textbf{66.92} & 36.85          & \textbf{16.39} & \textbf{84.20} & 95.07          & 23.85          \\ \midrule \midrule
\rowcolor[HTML]{D9F5D6} 
Qwen3-30B-A3B-Base   & 58.79          & 33.73          & 10.51          & 74.70          & 91.10          & 16.36          \\
+ Rand.                & 59.57          & 36.33          & 12.99          & 79.20          & 82.40          & 14.58          \\
+ R-Zero               & 62.38          & 34.69          & 11.00          & 82.20          & 94.31          & 15.60          \\
+ PretrainZero              & 64.59          & 36.58          & 14.91          & 81.70          & 94.40          & 17.40          \\ \midrule
+ ALIVE-Self           & 67.02          & 38.44          & \textbf{16.97} & 83.00          & 95.53          & \textbf{19.69} \\
+ ALIVE-External         & \textbf{67.15} & \textbf{39.20} & 16.79          & \textbf{84.50} & \textbf{95.83} & 19.38          \\ \bottomrule[1.2pt]
\end{tabular}%
}
\vspace{-4mm}
\end{table}

To assess the role of rich reviewer feedback in adversarial self-learning, we compare ALIVE with a spectrum of self-evolving reasoning baselines. \textbf{Rand.} uses random span masking as the simplest self-learning baseline. \textbf{R-Zero}~\citep{R-Zero} and \textbf{PretrainZero}~\citep{pretrainZero} are representative SOTA adversarial learning methods for improving LLM reasoning: R-Zero uses separate challenger and solver models for zero-data self-evolution, while PretrainZero learns to construct informative masked-span tasks from given documents. Their feedback remains scalar or verifiable, without rich reviews like instructive verbal critiques or process-level soft scores. Models were trained on Wikipedia across three scales: Qwen3-4B, 8B, and 30B-A3B~\citep{yang2025qwen3technicalreport}. Results in Table~\ref{tab:self_evolving_results} reveal three key patterns:

\textbf{(1) Adaptive adversarial learning outperforms random masking.} While random masking (Rand.) offers basic self-supervision, its gains are marginal and unstable, even degrading reasoning performance (e.g., GSM8K, AIME24) at the 30B-A3B scale. In contrast, ALIVE-Self consistently improves across all scales, proving that robust self-improvement requires adaptive task construction and reviewer-guided learning rather than naive masked prediction.

\textbf{(2) Dense reviewer feedback surpasses sparse-reward baselines.} Unlike R-Zero and PretrainZero which rely on sparse binary signals, ALIVE's Reviewer provides soft process-level scores and instructive critiques. This richer feedback drives consistent improvements: on Qwen3-8B, ALIVE-Self beats PretrainZero across all benchmarks, including \textbf{+4.06} points on AIME24, \textbf{+2.44} on SuperGPQA, and \textbf{+1.68} on BBEH. Similar gains hold at the 4B and 30B-A3B scales.

\textbf{(3) Self-review approaches powerful external reviewer supervision.} 
ALIVE-Self remains close to ALIVE-External across model scales. For Qwen3-8B, ALIVE-Self achieves 66.82\% on MMLU-Pro versus 66.92\% for ALIVE-External, and slightly surpasses it on SuperGPQA (36.90\% vs.~36.85\%) and AIME24 (24.06\% vs.~23.85\%). This indicates that, driven by continuous adversarial learning and hindsight-grounded feedback, internally generated critiques can provide training signals comparable to those from a powerful external reviewer.

\textbf{Takeaway.} 
\textit{Across adversarial self-learning baselines, ALIVE's advantage comes from enriching the learning signal: the Constructor creates challenging hindsight-verifiable tasks, while the Reviewer turns sparse outcomes into actionable verbal feedback. This enables more effective reasoning self-improvement than random masking, R-Zero, or PretrainZero under the same raw-corpus training setting.}

\subsection{RQ2: Scalability Without External Supervision}
\label{sec:exp-scalability}

Having established the effectiveness of rich feedback, we next investigate whether ALIVE can match supervised state-of-the-art methods without using ground-truth answers as direct training targets. We compare against GRPO~\citep{DeepSeekMath} with scalar outcome rewards, RFT~\citep{RFT} with supervised correct-solution tuning, FCP~\citep{FCP} with feedback-conditioned policy learning, and SDPO-Judge, an SDPO-style self-distillation baseline using Kimi-K2 to provide label-anchored diagnostic critiques. All methods share the same Qwen2.5-7B-Base backbone and training data (Big-Math~\citep{albalak2025bigmathlargescalehighqualitymath}, WebInstruct~\citep{ma2025generalreasoner}).

\begin{table}[th]
\centering
\footnotesize
\setstretch{1.2}
\caption{\textbf{Comparison of external feedback methods.} ALIVE surpasses both scalar-feedback (GRPO) and verbal-feedback (FCP) baselines.}
\label{tab:fcp_comparison}
\resizebox{0.7\textwidth}{!}{%
\begin{tabular}{lcccc}
\toprule[1.2pt]
\textbf{Method} & \textbf{GPQA-Diamond} & \textbf{MMLU-Pro} & \textbf{Math500} & \textbf{AIME24} \\
\midrule
\rowcolor[HTML]{D9F5D6}
Qwen2.5-7B-Base   & 27.90           & 49.70             & 63.80            & 7.50            \\ \midrule
+ GRPO            & 32.50           & 49.70             & 75.70            & 20.00           \\
+ RFT             & 35.20           & 55.00             & 69.20            & 13.30           \\
+ RFT + GRPO      & 37.20           & 57.00             & 75.10            & 25.80           \\
+ FCP             & 35.00           & 53.60             & 68.70            & 7.50            \\
+ FCP + Bootstrap & 39.10           & 55.30             & 76.50            & 25.00           \\
+ SDPO-Judge      & 40.90           & 55.60             & 77.40            & 26.00           \\ \midrule
+ ALIVE-Self      & \textbf{45.96}  & \textbf{61.32}    & \textbf{78.40}   & \textbf{26.67}  \\
+ ALIVE-External   & 44.95           & 61.16             & 77.60            & \textbf{26.67}  \\ \bottomrule[1.2pt]
\end{tabular}%
}
\end{table}

Table~\ref{tab:fcp_comparison} shows that ALIVE-Self achieves \textbf{strict Pareto dominance} across all four benchmarks:

\textbf{(1) Expert-level reasoning breakthrough.} On GPQA-Diamond, ALIVE-Self attains \textbf{45.96\%}, surpassing the strongest baseline (FCP+Bootstrap at 39.1\%) by \textbf{6.86 absolute points}, suggesting that ALIVE's self-supervised curriculum unlocks qualitatively different reasoning capabilities.

\textbf{(2) Consistent gains across domains.} The improvements are not limited to science: ALIVE-Self achieves \textbf{61.32\%} on MMLU-Pro (+6.02 over FCP+Bootstrap) and \textbf{78.40\%} on Math500 (+1.9 over SDPO-Judge). This demonstrates that the framework produces \textit{general} reasoning improvements rather than narrow task-specific tuning.

\textbf{(3) External supervision provides diminishing returns.} ALIVE-External achieves 44.95\% on GPQA-Diamond---1.01 points below ALIVE-Self. This inversion suggests that when the model is forced to generate its own critiques, it develops a deeper, more actionable understanding of its reasoning failures than when receiving external corrections.

\textbf{Takeaway.} \textit{ALIVE's self-supervised pipeline not only eliminates the need for ground-truth annotations but produces superior reasoning performance, particularly on expert-level tasks where external supervision is most expensive to obtain.}

\subsection{RQ3: Generalization to Complex Domains}
\label{sec:exp-generalization}

The previous results establish ALIVE's effectiveness on conventional reasoning benchmarks across general knowledge and mathematics. We now examine whether these gains transfer to more complex and less standardized domains: long-horizon coding tasks that require procedural consistency, and logical-boundary detection tasks that require recognizing whether the given information is sufficient for reasoning. We train {Qwen3-30B-A3B-Instruct} on coding corpora (SWE-smith, CodeContests, NuminaMath) and evaluate on LiveCodeBench~\citep{jain2024livecodebench}, SWE-bench Verified~\citep{jimenez2024swebenchlanguagemodelsresolve}, and QuestBench~\citep{li2025questbench}.

\begin{wrapfigure}{r}{0.45\textwidth}
\begin{minipage}{0.45\textwidth}
\centering
\footnotesize
\setstretch{1.1}
\vspace{-12mm}
\captionof{table}{
Performance on \textbf{Advanced Coding Benchmarks}. 
SWE-bench Verified uses the same mini-SWE-agent 
bash-only setup for all methods.
}
\label{tab:code_results}
\resizebox{\textwidth}{!}{%
\begin{tabular}{lcc}
\toprule[1.2pt]
\textbf{Method} & LiveCodeBench & SWE-bench Ver. \\
& (Pass@1) & (\% Resolved) \\
\midrule
\rowcolor[HTML]{FAF1D1}
Qwen3-30B-A3B-Instruct & 54.30 & 11.80 \\
+ SFT & 55.10 & 13.60 \\
+ GRPO & 55.40 & 14.80 \\
+ FCP & 54.90 & 14.00 \\
\midrule
+ ALIVE-Self & \textbf{56.00} & 17.20 \\
+ ALIVE-External & 55.80 & \textbf{17.60} \\
\bottomrule[1.2pt]
\end{tabular}%
}

\captionof{table}{
\textbf{Results on QuestBench.} 
}
\label{tab:questbench_results}
\resizebox{\textwidth}{!}{%
\begin{tabular}{@{}lcc@{}}
\toprule[1.2pt]
\textbf{Model} & \textbf{Logic-Q} & \textbf{Plan-Q} \\
\midrule
DeepSeek-V3.2 & 27.13\% & 23.65\% \\
GPT-4o & 32.78\% & 14.51\% \\
Kimi-K2 & 15.13\% & 21.03\% \\
GPT-5.5 & 46.52\% & 32.75\% \\
\midrule
Qwen3-30B-A3B-Instruct & 40.18\% & 8.50\% \\
+ ALIVE-Self & \textbf{43.91\%} & \textbf{31.35\%} \\
\bottomrule[1.2pt]
\end{tabular}%
}
\end{minipage}
\end{wrapfigure}

\subsubsection{Long-Horizon Agentic Coding}

\textbf{Results.} Table~\ref{tab:code_results} shows ALIVE-Self achieves \textbf{17.2\%} on SWE-bench Verified, a \textbf{+2.4\%} improvement over GRPO (14.8\%) and \textbf{+3.2\%} over FCP (14.0\%). Notably, the absolute gain on SWE-bench Verified (+5.4 points over base model) exceeds that on LiveCodeBench (+1.7 points), suggesting that \textit{rich reviewer feedback provides disproportionate benefits for harder, more open-ended tasks} where failure modes are complex and multi-factorial.

\textbf{Analysis.} The gains are modest on LiveCodeBench, whose tasks are relatively self-contained and evaluated by single-shot pass@1, but larger on SWE-bench Verified, where all methods operate under the same mini-SWE-agent bash-only setup and must maintain consistency across repository inspection, localization, patching, and validation. This gap suggests that ALIVE's construction-review loop is most useful for multi-step, process-dependent failures, where reviewer feedback supplies diagnostic signals beyond final patch success.

\subsubsection{Sensitivity to Logical Boundaries}

\textbf{Results.} Table~\ref{tab:questbench_results} shows that ALIVE improves information acquisition on both QuestBench domains: On Logic-Q, the gain is moderate (40.18\% $\rightarrow$ 43.91\%), while on Plan-Q---where the model must select the missing initial-state atom that disambiguates the shortest plan---ALIVE improves Qwen3-30B-A3B-Instruct by \textbf{+22.85 points} (8.50\% $\rightarrow$ 31.35\%), surpassing GPT-4o by \textbf{+16.84 points}.

\textbf{Analysis.} QuestBench evaluates whether a model can identify the variable needed to make an underspecified CSP solvable, rather than whether it can solve a fully specified problem. This aligns with ALIVE's training loop: the Constructor removes reasoning-critical information, and the Reviewer teaches the model which hidden span controls the correctness of the solution. The stronger Plan-Q gain suggests that this signal transfers especially well to procedural settings, where the key challenge is to locate the state information that determines downstream planning decisions.

\textbf{Takeaway.} \textit{ALIVE's construction-review loop improves not only answer generation but also information acquisition: the model becomes better at identifying which missing condition is necessary for reasoning, with the largest gains appearing in partially observed planning tasks.}

\subsection{RQ4: Ablation Study of Co-evolution Paradigm}
\label{sec:exp-ablation}
\subsubsection{Framework Ablations}
Table~\ref{tab:component_ablation} separates two kinds of interventions. The \textit{Only Constructor}, \textit{Only Solver}, and \textit{Only Reviewer} rows are \emph{decoupled role training}: we instantiate independent {Qwen3-8B-Base} models and optimize only the named role. In particular, \textit{Only Constructor} is trained only with the task-difficulty objective, \textit{Only Solver} is trained with the solving objectives (hard/soft rewards and FCP), and \textit{Only Reviewer} is trained only by critique distillation from Kimi-K2. Thus, \textit{Only Constructor} is a task-generation-only model and is not the same as training a Solver on constructor-generated queries. In contrast, the \textit{w/o} variants remove roles within the unified loop: \textit{w/o Constructor} replaces dynamic construction with a fixed Kimi-K2-generated QA pool, while \textit{w/o Reviewer} removes soft scoring and verbal feedback, leaving only sparse exact-match rewards for the Solver.

\begin{table}[ht]
\centering
\setstretch{1.1}
\footnotesize
\caption{Component-wise ablation of the Constructor, Solver, and Reviewer roles.}
\label{tab:component_ablation}
\resizebox{\textwidth}{!}{%
\begin{tabular}{lccccccccc}
\toprule[1.2pt]
\multicolumn{1}{c}{\textbf{Model}} & \textbf{Constructor} & \textbf{Solver} & \textbf{Reviewer} & \textbf{MMLU-Pro} & \textbf{SuperGPQA} & \textbf{BBEH}  & \textbf{Math500} & \textbf{GSM8K} & \textbf{AIME24} \\ \midrule
\rowcolor[HTML]{FAF1D1} 
\textbf{Qwen3-8B-Base} & \multicolumn{3}{l}{\cellcolor[HTML]{FAF1D1}}  & 59.15             & 31.12              & 10.49          & 70.10            & 91.50          & 10.00           \\ \midrule
Only Constructor & \cmark &        &        & 58.74 & 30.90 & 10.80 & 73.20 & 91.81 & 9.48  \\
Only Solver &        & \cmark &        & 61.62 & 33.97 & 14.58 & 78.90 & 92.49 & 19.89 \\
Only Reviewer &        &        & \cmark & 59.25 & 31.45 & 11.08 & 72.70 & 91.66 & 11.15 \\
w/o Constructor &        & \cmark & \cmark & 62.13 & 34.92 & 15.60 & 81.60 & 95.07 & 21.77 \\
w/o Reviewer & \cmark & \cmark &        & 64.20 & 34.34 & 14.40 & 80.10 & 93.25 & 20.83 \\
\textbf{ALIVE} & \cmark               & \cmark          & \cmark            & \textbf{66.82}   & \textbf{36.90}     & \textbf{16.35} & \textbf{83.80}   & \textbf{96.06} & \textbf{24.06}  \\ \bottomrule[1.2pt]
\end{tabular}%
}
\end{table}

Results show that:
\textbf{(1) Isolated roles are ineffective.} Only Constructor (58.74\%) and Only Reviewer (59.25\%) are comparable to or slightly below the base model (59.15\%), while Only Solver improves to 61.62\% but remains 5.2 points behind full ALIVE, indicating that no single role yields substantial gains in isolation.
\textbf{(2) The Constructor and Reviewer provide synergistic benefits.} Both w/o Constructor and w/o Reviewer degrade performance, confirming that Constructor and Reviewer provide complementary and synergistic benefits.

\subsubsection{Constructor-Solver Coupling}
\begin{table}[t]
\centering
\footnotesize
\setstretch{1.1}
\caption{Effect of Constructor--Solver co-evolution. 
S: shared parameters with Solver; 
T: trainable Constructor; 
E: external Constructor (off-the-shelf LLM). }
\label{tab:constructor_independence}
\small
\setlength{\tabcolsep}{4pt}
\begin{tabular}{lccccccccc}
\toprule[1.2pt]
\textbf{Method} 
& \textbf{S} 
& \textbf{T} 
& \textbf{E}
& \textbf{MMLU-Pro} & \textbf{SuperGPQA} & \textbf{BBEH} & \textbf{Math500} & \textbf{GSM8K} & \textbf{AIME24} \\
\midrule
Frozen        & \xmark & \xmark & \xmark & 60.25 & 33.87 & 13.30 & 74.60 & 92.42 & 19.90 \\
Decoupled     & \xmark & \cmark & \xmark & 63.35 & 34.93 & 15.58 & 80.10 & 93.78 & 20.52 \\
External      & \xmark & \xmark & \cmark & 62.58 & 34.95 & 15.60 & 81.90 & 95.22 & 21.77 \\ \midrule
\textbf{ALIVE}& \textbf{\cmark} & \textbf{\cmark} & \textbf{\xmark}
              & \textbf{66.82} & \textbf{36.90} & \textbf{16.35}
              & \textbf{83.80} & \textbf{96.06} & \textbf{24.06} \\
\bottomrule[1.2pt]
\vspace{-4mm}
\end{tabular}
\end{table}
This section examines whether dynamic task construction alone is sufficient or whether the Constructor must co-evolve with the Solver within a unified policy. We compare 4 settings: (1) \textbf{Frozen}, which uses a fixed \texttt{Qwen3-8B-Base} Constructor; (2) \textbf{Decoupled}, where the Constructor is trainable with the same difficulty reward but optimized independently from the Solver without parameter sharing; (3) \textbf{External}, which replaces the Constructor with a stronger off-the-shelf LLM (Kimi-K2.5); and (4) the full \textbf{ALIVE} setting, where Constructor and Solver share parameters and co-evolve within a single policy.

Results in Table~\ref{tab:constructor_independence} show that: 
\textbf{(1) Frozen Constructors fail.} A fixed Constructor achieves 60.25\% on MMLU-Pro, only marginally above the base model, indicating that effective task generation must adapt to the Solver.  
\textbf{(2) Decoupling limits performance.} Independently trained Constructor improves to 63.35\% but remains below ALIVE, suggesting that lack of shared optimization prevents proper difficulty alignment.  
\textbf{(3) Stronger external Constructors are insufficient.} Even replacing the Constructor with Kimi-K2.5 (a substantially stronger model) achieves only 62.58\%---\textbf{4.2 points below} the unified system. This is the most striking finding: a stronger but decoupled Constructor performs worse than a weaker but co-evolving one.

\textbf{Takeaway.} \textit{ALIVE's efficacy stems not from any individual component but from the unified co-evolution of all three roles. Parameter sharing enables an adaptive curriculum that external or decoupled alternatives cannot replicate, even with superior base capabilities.}

\section{Related Work}
\label{sec:related_work_main}

\textbf{Reinforcement Active Learning.}
Traditional active learning uses heuristics to select data for labeling. Reinforcement Active Learning instead learns a data-selection policy through RL rewards, optimizing long-term model improvement. R-Zero\citep{R-Zero} and Pretrain-Zero\citep{pretrainZero} advance this by eliminating human annotation via adversarial data generation, but rely on scalar rewards that ignore reasoning quality. Our work addresses this gap by introducing reasoning-aware feedback into the adversarial curriculum.

\textbf{Learning from Verbal Feedback.}
To provide richer supervision than scalar rewards, recent work explores learning from natural language feedback, critiques, or process-level explanations \citep{stephan2024rlvflearningverbalfeedback, naturallanguageactorcritic, Critique-GRPO, FCP}. These approaches leverage textual diagnostics to guide reasoning, typically by conditioning generation on externally provided or pre-generated feedback.

Unlike prior methods, ALIVE unifies task construction, problem-solving, and self-evaluation into a unified self-supervised reinforcement learning loop, enabling models to generate and learn from their own reasoning-rich verbal feedback directly from raw text, without external reward annotations.

\section{Conclusion}
\label{sec:discussion}
This paper introduced ALIVE, a self-supervised framework that rethinks reasoning acquisition as a closed-loop, self-evolving process. ALIVE integrates task construction, problem-solving, and self-review into a single policy, using self-generated verbal critiques rather than external rewards. Experiments across general-domain reasoning, mathematical problem solving, and coding demonstrate that ALIVE improves reasoning accuracy, cross-domain generalization, and emergent logical gap detection. This work establishes a scalable framework in which reasoning ability emerges from structured self-play rather than from costly external supervision.

{\small
\bibliographystyle{plainnat}
\bibliography{references}
}
\clearpage
\appendix

\section{Related Work}
\label{sec:related_work}

This appendix provides a broader positioning of ALIVE relative to prior reasoning-alignment paradigms. Table~\ref{tab:method-comparison} summarizes the main differences in terms of dependence on human-written reward annotations, explicit environments or verifiers, domain-specific design, and the type of feedback signal used for optimization.

\begin{table}[ht]
\small
    \caption{Comparison of training paradigms. ``Reward-ann. free'' indicates that the method does not require human-written reward annotations.}
    \label{tab:method-comparison}
    \begin{center}
    \begin{tabular}{lccccc}
        \toprule[1.2pt]
        \multirow{2}{*}{\textbf{Method}} & \textbf{Reward Ann.} & \textbf{Environment} & \textbf{Domain} & \multicolumn{2}{c}{\textbf{Reward Signal}} \\
        \cmidrule(lr){5-6}
         & \textbf{Free} & \textbf{Free} & \textbf{General} & \textbf{Scalar} & \textbf{Verbal} \\
        \midrule
        RL(H/AI)F  & \xmark & \cmark & \xmark & \cmark & \xmark \\
        RLVR  & \cmark & \xmark & \xmark & \cmark & \xmark \\
        PretrainZero & \cmark & \cmark & \cmark & \cmark & \xmark \\
        FCP   & \cmark & \cmark & \xmark & \xmark & \cmark \\
        \midrule
        \rowcolor{gray!10} 
        \textbf{ALIVE (Ours)} & \cmark & \cmark & \cmark & \cmark & \cmark \\
        \bottomrule[1.2pt]
    \end{tabular}
    \end{center}
\end{table}

\paragraph{Self-evolving Task Generation.}
A first line of work reduces reliance on manually collected training tasks by letting models generate or transform their own data. R-Zero~\citep{R-Zero} and PretrainZero~\citep{pretrainZero} use generator-solver style training to construct increasingly useful reasoning data, while Writing-Zero~\citep{Writing-Zero} extends self-evolving training toward less directly verifiable generation tasks. These methods address the data bottleneck, but their optimization is still primarily driven by scalar or verifiable rewards; failures usually remain low-bandwidth signals rather than instructive critiques.

\paragraph{Learning from Richer Feedback.}
A second line increases the information bandwidth of feedback. RL(H/AI)F~\citep{RLHF,RLAIF,rlhfbook} uses human or AI preferences, but typically requires task-specific feedback collection. RLVR~\citep{rlveriabler} reduces human reward annotation by relying on verifiable environments, yet it is constrained by the availability of reliable checkers. Natural-language feedback and actor-critic variants~\citep{stephan2024rlvflearningverbalfeedback,naturallanguageactorcritic,Text2Grad,rapo}, feedback-conditioned policies~\citep{FCP}, self-distillation from feedback~\citep{SDPO}, and generative or critique-based verifiers~\citep{SCoRe,GenerativeVerifiers,Critique-GRPO} explore ways to make feedback more informative than a single binary outcome.

\paragraph{Positioning of ALIVE.}
ALIVE combines these two directions in a single parameter-shared loop: the model constructs reasoning-critical tasks, solves them, and reviews its own reasoning with both hard verification and instructive verbal feedback. Compared with self-evolving data-generation methods, ALIVE adds a Reviewer that turns failures into dense diagnostic signals. Compared with rich-feedback methods, ALIVE does not rely on a fixed feedback source alone; feedback is coupled to an adaptive Constructor that tracks the Solver's capability frontier. As summarized in Table~\ref{tab:method-comparison}, ALIVE is designed to reduce dependence on human-written reward annotations while remaining environment-free, domain-general, and compatible with both scalar and verbal reward signals.

\section{Benchmark}
To comprehensively evaluate model performance across diverse cognitive dimensions, we utilize a suite of benchmarks spanning general knowledge, mathematics, programming, and complex reasoning.

\subsection{General Domain}
\paragraph{MMLU-Pro.} 
MMLU-Pro \citep{wang2024mmlu} is an enhanced version of the Massive Multitask Language Understanding (MMLU) benchmark, developed to address the saturation of performance on general knowledge tasks. It expands both the number of answer choices per question and the complexity of reasoning across 14 diverse subjects. By lowering the probability of success through random guessing, MMLU-Pro provides a more rigorous evaluation of a model’s academic knowledge and reasoning abilities.

\paragraph{GPQA-Diamond.} 
GPQA-Diamond~\citep{GPQA} is a carefully curated subset of the GPQA benchmark, consisting of 198 multiple-choice questions in biology, chemistry, and physics. The questions span advanced undergraduate to postgraduate difficulty levels. Each item was selected such that domain experts answered correctly while the majority of non-experts did not, ensuring both high discriminative power and quality.

\paragraph{SuperGPQA.} 
SuperGPQA~\citep{pteam2025supergpqascalingllmevaluation} is a high-difficulty benchmark designed to evaluate expert-level knowledge across a wide range of scientific and technical domains. Building on GPQA’s methodology, it spans a broader set of specialized subjects and demands deep domain expertise. The tasks are crafted to challenge even human experts, offering a rigorous assessment of large language models’ capabilities in advanced scientific reasoning.

\paragraph{BBEH.} 
BBEH (Big-Bench Esports Hard)~\citep{kazemi-etal-2025-big} is a distilled subset of BIG-bench, focusing on tasks where language models historically fail to reach human-level performance. It spans linguistic, logical, and creative challenges requiring nuanced understanding and multi-step reasoning. BBEH serves as a critical measure of a model’s capacity to tackle “long-tail” reasoning problems beyond patterns typically found in training corpora.

\subsection{Math Domain}
\paragraph{MATH-500.} 
MATH-500~\citep{hendrycks2021measuring} evaluates mathematical reasoning and problem-solving abilities, addressing the need for challenging assessments as model capabilities grow. It contains 500 problems across five domains: algebra, combinatorics, geometry, number theory, and precalculus. Each problem requires multi-step reasoning and advanced problem-solving, beyond simple calculations or factual recall.

\paragraph{GSM8K.} 
GSM8K~\citep{cobbe2021training} is a collection of grade-school math word problems designed to assess multi-step arithmetic and quantitative reasoning. Problems typically involve two to eight reasoning steps and linguistic diversity, making GSM8K a standard benchmark for chain-of-thought evaluation and verifier-based training.

\paragraph{AIME24.} 
AIME24~\citep{aime24} contains 30 problems from the 2024 American Invitational Mathematics Examination (AIME), a prestigious high school competition renowned for its challenging questions.

\subsection{Code Domain}
\paragraph{LiveCodeBench.} 
LiveCodeBench~\citep{jain2024livecodebench} evaluates real-time code generation and editing by language models. It includes diverse programming tasks across multiple languages and emphasizes interactive scenarios such as incremental edits, debugging, and code completion, reflecting practical coding workflows.

\paragraph{SWE-bench Verified.} 
SWE-bench Verified~\citep{jimenez2024swebenchlanguagemodelsresolve} is a human-validated subset of the SWE-bench software-engineering benchmark. It focuses on real-world GitHub issues and patch generation, providing a reliable evaluation of a model’s ability to propose correct, test-passing fixes and produce high-quality software under realistic conditions.

\subsection{Logic Reasoning Domain}

QuestBench \citep{li2025questbench} formalizes the problem of information gathering as an underspecified Constraint Satisfaction Problem (CSP). Specifically, it focuses on identifying a ``1-sufficient'' variable, where the value of a target variable $y$ cannot be inferred from the given information alone.

\paragraph{Logic-Q.}
This dataset consists of propositional logic tasks adapted from the SimpleLogic benchmark. Each problem presents a set of implicative rules (e.g., ``If Alice is A and B, then Alice is C'') and a set of known properties. The problem is under-specified such that the truth value of the target conclusion depends on exactly one missing proposition about the subject. The model must identify which attribute to query to resolve the ambiguity.

\paragraph{Planning-Q.}
Planning-Q is based on the Blocks World domain within the Planning Domain Definition Language (PDDL). These tasks involve finding the shortest sequence of actions to rearrange blocks from an initial state to a goal state. The initial state is only partially observed (e.g., the position of a specific block is unknown), leading to multiple optimal plans. The model must ask for the specific state atom that disambiguates the shortest path to the goal.

\section{Implementation Details}
\label{app:training_details}

\subsection{Training Datasets}
\label{app:train_data}
We train the model using a mixture of publicly available datasets, including \textbf{Wikipedia}\footnote{\url{https://huggingface.co/datasets/wikimedia/wikipedia}}, \textbf{Big-Math}\footnote{\url{https://huggingface.co/datasets/SynthLabsAI/Big-Math-RL-Verified}}, \textbf{WebInstruct}\footnote{\url{https://huggingface.co/datasets/TIGER-Lab/WebInstruct-verified}}, \textbf{SWE-smith}\footnote{\url{https://huggingface.co/datasets/SWE-bench/SWE-smith}}, \textbf{CodeContests}\footnote{\url{https://huggingface.co/datasets/deepmind/code_contests}}, and \textbf{NuminaMath-CoT}\footnote{\url{https://huggingface.co/datasets/AI-MO/NuminaMath-CoT}}. These datasets span diverse domains, including general knowledge, mathematical reasoning, code generation, and instruction-following, enabling balanced capability acquisition across reasoning and generation tasks.

\subsection{Experiments Environment}
All training and inference are conducted on a cluster of 32 NVIDIA H20 (96GB) GPUs. We use the VeRL framework \citep{verl} for reinforcement learning optimization and vLLM \citep{vllm} for high-throughput inference. The full training process consists of 2,048 optimization steps and requires approximately 70 hours under our default configuration.

\subsection{Optimization Objective}
We optimize a unified objective that jointly models curriculum construction, reasoning optimization, and feedback-based internalization. The overall objective is defined as:
\begin{equation}
\mathcal{J}_{\text{Total}}(\theta) = \underbrace{\mathcal{J}_{\text{GRPO}}^{\text{Const}}(\theta)}_{\text{Curriculum}} + \underbrace{\mathcal{J}_{\text{GRPO}}^{\text{Solver}}(\theta)}_{\text{Reasoning}} - \lambda_2 \underbrace{\mathcal{L}_{\text{FCP}}^{\text{Reviewer}}(\theta)}_{\text{Self-correction}} - \lambda_3 \underbrace{\mathcal{L}_{\text{distill}}(\theta)}_{\text{Reviewer Warm-up}},
\end{equation}
where $\mathcal{J}_{\text{Const}}$ and $\mathcal{J}_{\text{Solver}}$ are policy gradient objectives optimized via GRPO, corresponding respectively to curriculum construction and reasoning generation. The term $\mathcal{L}_{\text{FCP}}$ is a negative log-likelihood loss that encourages feedback-conditioned self-correction, while $\mathcal{L}_{\text{distill}}$ corresponds to supervised distillation of reviewer behavior and is only active during the warm-up stage. Importantly, the distillation loss is not applied during reinforcement learning updates in the self-play phase, ensuring that policy optimization is driven solely by RL signals after initialization.

\subsection{Training Pipeline}
We adopt a two-stage training pipeline consisting of a Critique Distillation Warm-up phase followed by an Autonomous Self-Play Reinforcement Learning phase. Both stages operate under the same trajectory generation mechanism, ensuring consistency in data distribution across training phases.

Each training iteration is based on a fixed rollout configuration. Specifically, for each input, we first sample $M=8$ constructor outputs, and for each constructor output, we further generate $N=16$ solver rollouts. This yields a total of 128 trajectories per input per iteration. In this work, a trajectory refers to a complete interaction tuple consisting of the constructor output, solver solution, and feedback signal, which serves as the basic unit for reinforcement learning optimization. Accordingly, all reported batch sizes correspond to the number of trajectories used per parameter update.

\paragraph{Critique Distillation Warm-up}
We first perform a warm-up phase of 256 training steps to initialize the reviewer capability. During this stage, the model learns to generate critiques from an external teacher model (Kimi-K2). For each training sample $(\tilde{x}, \hat{y}, y^*)$, we optimize a supervised distillation objective:
\begin{equation}
    \mathcal{L}_{\text{distill}} = - \log \pi_\theta(c_{\text{teacher}} \mid \tilde{x}, \hat{y}, y^*).
\end{equation}
This loss is applied exclusively during the warm-up stage and does not interact with reinforcement learning objectives. We set the distillation coefficient $\lambda_3 = 1.0$ during this phase to fully align the reviewer behavior with the teacher model before RL-based self-improvement begins.

\paragraph{Autonomous Self-Play Phase}
After warm-up, we remove access to the external teacher model and transition to fully autonomous self-play. In this stage, the model serves simultaneously as constructor, solver, and reviewer, enabling end-to-end self-improvement through reinforcement learning. The distillation objective is disabled by setting $\lambda_3 = 0$, and training continues for the remaining steps up to 2,048 total iterations.

\subsection{Hyperparameters}
\label{app:Hyperparameter}
We optimize all models using the AdamW optimizer with a learning rate of $7.5 \times 10^{-7}$. For each input, we use the same rollout configuration described above ($M=8$, $N=16$) with a sampling temperature of 1.0 to ensure consistent exploration across all training stages.

To stabilize policy optimization, we apply PPO-style trust-region clipping to the probability ratio. For base models trained with GRPO, we use a clipping range of $\epsilon \in [0.2, 0.28]$. For the large-scale MoE model (Qwen3-30B-A3B) trained under GSPO, we adopt a significantly tighter range of $\epsilon \in [3 \times 10^{-4}, 4 \times 10^{-4}]$, which we find necessary to mitigate instability caused by sparse expert routing and higher variance in policy updates.

The overall objective includes two weighting coefficients that control the balance of learning signals. The feedback-conditioned policy loss is weighted by a fixed coefficient $\lambda_2 = 0.5$ throughout training to ensure stable internalization of critique signals.

For soft reward supervision, we introduce an adaptive coefficient $\lambda_1$ that depends on the length of the hindsight ground-truth $y$. Specifically, we set $\lambda_1 = 1.0$ when $\mathrm{len}(y) \geq 16$, and $\lambda_1 = 0.6$ otherwise. This design is motivated by the observation that shorter targets typically correspond to deterministic entities where exact matching is sufficient, whereas longer targets often encode compositional reasoning structures where soft semantic supervision provides a more informative learning signal than strict token-level matching.

\section{Additional Experiments}
\label{app:additional experiment}


\subsection{Interaction with DAPO-style filtering}

Table~\ref{tab:dapo_backend} evaluates whether ALIVE's adaptive construction can stack with a stronger RL backend. We replace the GRPO backend with DAPO-style filtering while keeping the ALIVE loop unchanged. This comparison is intended to test complementarity rather than to claim that DAPO and ALIVE solve the same problem: DAPO filters or reweights samples during optimization, whereas ALIVE changes which tasks are generated and how failures are reviewed.

\begin{table}[ht]
\centering
\caption{ALIVE with different policy optimization backends. DAPO-style filtering is complementary to ALIVE's adaptive task generation.}
\label{tab:dapo_backend}
\small
\begin{tabular}{lcccccc}
\toprule[1.2pt]
\textbf{Optimization Backend} & MMLU-Pro & SuperGPQA & BBEH & Math500 & GSM8K & AIME24 \\
\midrule
ALIVE-GRPO & 66.82 & 36.90 & \textbf{16.35} & 83.80 & 96.06 & 24.06 \\
ALIVE-DAPO & \textbf{66.86} & \textbf{36.94} & 16.33 & \textbf{84.00} & \textbf{96.13} & \textbf{24.27} \\
\bottomrule[1.2pt]
\end{tabular}
\end{table}

The results clarify the role interaction behind ALIVE. First, the Solver objective is the largest single source of improvement, but neither the Constructor nor the Reviewer is dispensable: removing either role consistently lowers performance relative to the full loop. Second, task construction alone does not automatically translate into reasoning gains unless it is coupled to the Solver; even a stronger frozen Constructor falls behind the unified co-evolving policy. Third, DAPO-style filtering brings a small additional gain when used as ALIVE's backend, indicating that optimizer-level filtering and ALIVE's construction-review loop are complementary rather than mutually exclusive.

\subsection{Experiment Observations}
\label{app:curve}

\begin{figure*}[t]
    \centering
    \begin{subfigure}[b]{0.32\textwidth}
        \centering
        \includegraphics[width=\textwidth]{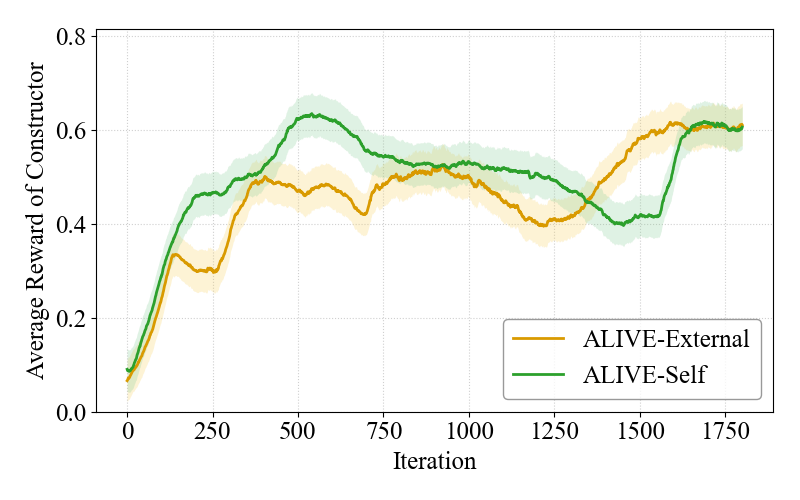}
        \caption{Average reward of Constructor}
        \label{fig:curve_reward}
    \end{subfigure}
    \hfill
    \begin{subfigure}[b]{0.32\textwidth}
        \centering
        \includegraphics[width=\textwidth]{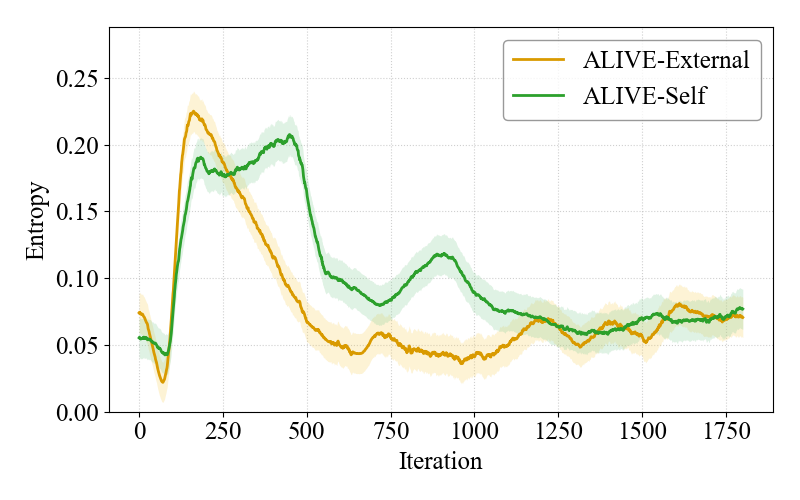}
        \caption{Policy entropy}
        \label{fig:curve_entropy}
    \end{subfigure}
    \hfill
    \begin{subfigure}[b]{0.32\textwidth}
        \centering
        \includegraphics[width=\textwidth]{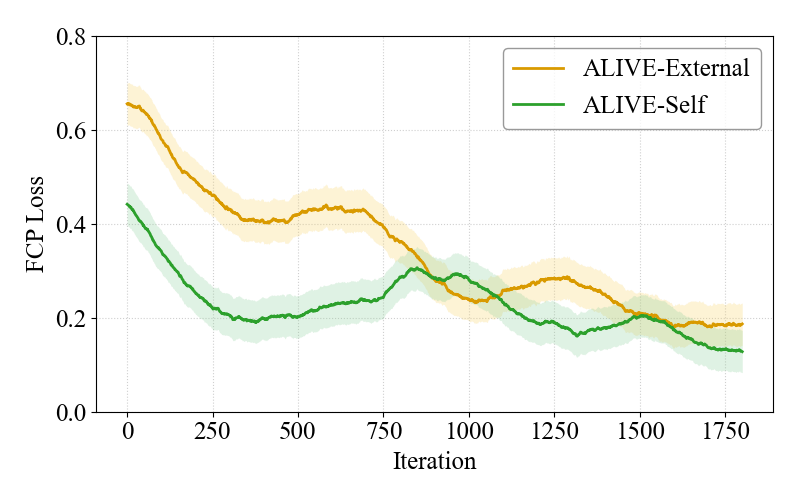} 
        \caption{FCP loss}
        \label{fig:curve_loss}
    \end{subfigure}
    \caption{\textbf{Training dynamics.} A comparison of the Constructor's reward, policy entropy, and FCP loss between the fully autonomous ALIVE-Self (red) and the External-guided ALIVE-External (black).}
    \label{fig:training_curves}
\end{figure*}

We analyze the training dynamics of \textbf{ALIVE-Self} and \textbf{ALIVE-External} using three representative metrics, as illustrated in Figure~\ref{fig:training_curves}.

\paragraph{Adversarial Co-Evolution.}
Figure~\ref{fig:curve_reward} reports the \textbf{Constructor Reward} over training, showing distinct evolutionary behaviors. \textbf{ALIVE-Self} (red) exhibits a sharp early increase, peaking around step 500, suggesting that without external constraints, the Constructor rapidly identifies adversarial masking patterns that challenge the initially undertrained Solver. The subsequent decline between steps 500 and 1400 indicates that the Solver has adapted to these initial strategies, reducing the Constructor’s reward. After step 1500, the reward rises again as the Constructor develops more complex masking strategies, reflecting ongoing adaptation between the two roles.
In contrast, \textbf{ALIVE-External} (black) follows a smoother, more monotonic trajectory, consistent with a stable but less explorative training curriculum.

\paragraph{Exploration vs. Exploitation.}
Figure~\ref{fig:curve_entropy} shows the policy entropy over training. \textbf{ALIVE-Self} maintains substantially higher entropy during the early phase (steps 200–500) and exhibits a secondary spike around step 900, coinciding with the drop in reward. This pattern indicates that the autonomous model engages in broader exploration to escape local optima. In contrast, the External-guided model collapses its search space earlier, driven by the strong supervisory signal from Kimi-K2, resulting in reduced exploratory behavior.

\paragraph{Alignment Efficiency.}
Figure~\ref{fig:curve_loss} shows that \textbf{ALIVE-Self} reduces the FCP Loss notably faster than the External baseline during the early stages. This suggests that the model can internalize and predict its \textit{own} critique logic (self-consistency) more efficiently than aligning with an external teacher’s distribution. The temporary increase in loss around step 800 corresponds to the Solver adapting to new Constructor strategies, which momentarily disrupts the previously learned critique patterns.

\section{Limitations and Broader Impacts}
\label{app:limitations}

ALIVE reduces dependence on human-written reward annotations, but it is not free of assumptions or external anchors. The Reviewer warm-up uses a teacher model only to bootstrap critique format and calibration, and QA-derived corpora may contain ground-truth answers inside the raw document representation used for controlled comparisons. To avoid treating these labels as direct supervised targets, the Constructor may mask any reasoning-critical span; empirically, final answers account for only about $50$--$60\%$ of selected masks, with the remainder covering problem conditions, intermediate reasoning steps, or latent pivots. The method is therefore best suited to settings where a hidden target can be recovered or verified in hindsight; fully open-ended tasks may require additional human or tool-based validation.

The framework also increases compute relative to single-role RL because each document produces multiple constructed tasks and each task requires grouped Solver rollouts. In addition, self-generated critiques can inherit biases, factual errors, or unsafe content from the base model or source corpora. We mitigate this risk through hindsight ground truth, hard exact-match rewards, bounded soft rewards, and role-level ablations, but deployment would still require monitoring of generated tasks and critiques, especially on noisy or sensitive corpora. Positively, ALIVE can reduce the need for costly human reward annotation and improve self-correction in reasoning, mathematics, and software-engineering settings. Negatively, stronger reasoning and coding abilities may also improve harmful automation or unreliable decision support if used without domain-specific safeguards.

\newpage
\section{Prompt Templates}
\label{app:prompts}

\begin{tcolorbox}[
    breakable,
    colback=gray!5!white,
    colframe=gray!75!black,
    title=\textbf{Prompt of the Constructor}
]

\textbf{Role Description.}  
You are an expert \emph{Task Constructor} specializing in the design of reasoning-intensive evaluation tasks. Your purpose is to probe the depth and robustness of an advanced AI model’s reasoning ability.

\medskip
\textbf{Primary Objective.}  
Given a raw input document, construct a \textbf{reasoning-critical reconstruction task} (e.g., fill-in-the-blank or partial derivation) that cannot be solved through surface pattern matching and instead requires multi-step logical inference.

\medskip
\textbf{Construction Procedure.}
\begin{enumerate}
    \item \textbf{Identify the Logical Pivot.}  
    Locate the central reasoning component of the document—such as a key assumption, intermediate lemma, transformation, or decision point—that is necessary to derive the final conclusion.
    
    \item \textbf{Remove or Mask the Pivot.}  
    Elide this critical component or its derivation, while preserving the surrounding context.
    
    \item \textbf{Formulate the Task.}  
    Design a question that requires the solver to reconstruct the missing logic using only the remaining information.
\end{enumerate}

\medskip
\textbf{Design Constraints.}
\begin{itemize}
    \item Do \emph{not} mask arbitrary tokens or superficial details; always mask \textbf{reasoning-critical logic}.
    \item Ensure the retained context is sufficient for deduction in principle, but requires non-trivial, multi-step reasoning.
    \item The task should admit a clear, well-defined ground truth derivable from the original document.
\end{itemize}

\medskip
\textbf{Input Document.}
\begin{itemize}
    \item \textbf{Raw Text:} \texttt{\{\{RAW\_DOCUMENT\}\}}
\end{itemize}

\medskip
\textbf{Required Output Format.}

\texttt{\textless Thought\textgreater} 

A brief analysis identifying the logical pivot and justifying why it is the most challenging component to reconstruct. 

\texttt{\textless /Thought\textgreater}
    
\texttt{\textless Task\textgreater} 

The constructed question or partial context is presented to the Solver, with the critical gap clearly defined.  

\texttt{\textless /Task\textgreater}
    
\texttt{\textless Hidden\_Truth\textgreater} 

The exact content that was removed or masked is to be used exclusively by the Reviewer as external information.  

\texttt{\textless /Hidden\_Truth\textgreater}
\end{tcolorbox}

\begin{tcolorbox}[
    breakable,
    colback=gray!5!white,
    colframe=gray!75!black,
    title=\textbf{Prompt of the Solver}
]

\textbf{Role Description.}  
You are a precise and disciplined \emph{Task Solver}. Your objective is to solve the given problem correctly by applying structured, logically sound reasoning.

\medskip
\textbf{Primary Objective.}  
Derive the correct solution to the user’s task through a coherent sequence of logical steps, ensuring internal consistency and factual correctness.

\medskip
\textbf{Reasoning Guidelines.}
\begin{enumerate}
    \item Decompose the problem into minimal, well-defined subproblems.
    \item Explicitly state any assumptions or constraints used in the reasoning.
    \item For mathematical or algorithmic tasks, carry out intermediate checks to validate correctness.
    \item Ensure that each step follows logically from previous steps without gaps or unjustified leaps.
\end{enumerate}

\medskip
\textbf{Task Input.}
\begin{itemize}
    \item \textbf{User Query:} \texttt{\{\{CONSTRUCTED\_TASK\}\}}
\end{itemize}

\medskip
\textbf{Required Output Format.}

\texttt{\textless Reasoning\textgreater}

A complete, step-by-step derivation leading to the solution.  

\texttt{\textless /Reasoning\textgreater}

\texttt{\textless Answer\textgreater} 

The final, concise result is derived from the reasoning above. 

\texttt{\textless /Answer\textgreater}

\end{tcolorbox}

\begin{tcolorbox}[
    breakable,
    colback=gray!5!white,
    colframe=gray!75!black,
    title=\textbf{Prompt of the Reviewer}
]

\textbf{Role Description.}  
You are a rigorous and insightful \emph{Reasoning Reviewer}. Your task is to critically evaluate a solution produced by an AI \emph{Solver}, focusing on both \textbf{final correctness} and \textbf{reasoning validity}.

\medskip
\textbf{Provided Context.}
\begin{itemize}
    \item \textbf{Task Specification:} \texttt{\{\{CONSTRUCTED\_TASK\}\}}
    \item \textbf{Solver Output:} \texttt{\{\{SOLVER\_OUTPUT\}\}}
    \item \textbf{Hindsight Ground Truth:} \texttt{\{\{HIDDEN\_TRUTH\}\}}
\end{itemize}

\medskip
\textbf{Important Note.}  
The Solver did \emph{not} have access to the Hindsight Ground Truth. You may freely use it to assess correctness and diagnose reasoning errors.

\medskip
\textbf{Evaluation Procedure.}
\begin{enumerate}
    \item \textbf{Outcome Verification.}  
    Determine whether the Solver’s final answer aligns with the essential conclusion of the Hidden Truth.
    
    \item \textbf{Reasoning Analysis.}  
    Assess whether the Solver’s reasoning steps are logically valid and well-founded.  
    If the final answer is incorrect, identify the \emph{precise step} where the reasoning diverges from the correct logic.
    
    \item \textbf{Constructive Critique.}  
    Provide concise, actionable feedback explaining \emph{why} the reasoning is flawed or confirming why it is correct. Avoid vague judgments such as “incorrect” without justification.
\end{enumerate}

\medskip
\textbf{Required Output Format.}

\texttt{\textless Analysis\textgreater}

A step-by-step comparison between the Solver’s reasoning and the Hidden Truth.  

\texttt{\textless /Analysis\textgreater}
    
\texttt{\textless Critique\textgreater} 

A brief, instructive summary highlighting the core error or validating the reasoning. 

\texttt{\textless /Critique\textgreater}
    
\texttt{\textless Score\textgreater}  

A continuous score $v \in [0, 1]$ reflecting overall logical correctness.  

\texttt{\textless /Score\textgreater}
\end{tcolorbox}

\end{document}